\pdfoutput=1
\documentclass[11pt]{article}
\usepackage{EMNLP2023}

\usepackage{times}
\usepackage{latexsym}

\usepackage[T1]{fontenc}
\usepackage[utf8]{inputenc}

\usepackage{microtype}

\usepackage{microtype}
\usepackage{amsmath}

\usepackage{graphicx} 
\usepackage{float} 
\usepackage{subfigure} 
\usepackage{makecell}
\usepackage{epstopdf}
\usepackage{hyperref}
\usepackage{multirow}
\usepackage{multicol}
\usepackage{amsmath}
\usepackage{fancyhdr}
\usepackage{lipsum} 
\usepackage{inconsolata}

\title{Oasis: Data Curation and Assessment System for Pretraining of Large Language Models}

\author{
Tong Zhou\textsuperscript{1},
Yubo Chen\textsuperscript{1,2},
Pengfei Cao\textsuperscript{1,2},
Kang Liu\textsuperscript{1,2}, 
Jun Zhao\textsuperscript{1,2}, 
Shengping Liu\textsuperscript{3}
\\
 \textsuperscript{1} The Laboratory of Cognition and Decision Intelligence for Complex Systems \\ Institute of Automation, Chinese Academy of Sciences \\
 \textsuperscript{2} School of Artificial Intelligence, University of Chinese Academy of Sciences \\
 \textsuperscript{3} Beijing Unisound Information Technology Co., Ltd\\
  \texttt{tong.zhou@ia.ac.cn}\\
  \texttt{\{yubo.chen,pengfei.cao,kliu,jzhao\}@nlpr.ia.ac.cn}
  \texttt{liushengping@unisound.com}
}

\begin{document}
\maketitle

\begin{abstract}

Data is one of the most critical elements in building a large language model.
However, existing systems either fail to customize a corpus curation pipeline or neglect to leverage comprehensive corpus assessment for iterative optimization of the curation.
To this end, we present a pretraining corpus curation and assessment platform called Oasis\footnotemark[1]\footnotemark[2] -- a one-stop system for data quality improvement and quantification with user-friendly interactive interfaces. 
Specifically, the interactive modular rule filter module can devise customized rules according to explicit feedback. 
The debiased neural filter module builds the quality classification dataset in a negative-centric manner to remove the undesired bias. 
The adaptive document deduplication module could execute large-scale deduplication with limited memory resources. 
These three parts constitute the customized data curation module.
And in the holistic data assessment module, a corpus can be assessed in local and global views, with three evaluation means including human, GPT-4, and heuristic metrics.
We exhibit a complete process to use Oasis for the curation and assessment of pretraining data.
In addition, an 800GB bilingual corpus curated by Oasis is publicly released\footnotemark[2].

\end{abstract}

\begingroup
\setcounter{footnote}{0}% Reset footnote counter
\footnotetext[1]{Project:https://github.com/tongzhou21/Oasis} 
\footnotetext[2]{Video:https://youtu.be/YLfMlnrUZPk} 

\footnotetext[3]{Corpus: https://huggingface.co/datasets/Oasis-Team/Oasis-Corpus} 
\endgroup

\section{Introduction}

Building large language models (LLMs) for proficiency in versatility tasks has been spotlighted recently \citep{openai2023gpt,touvron2023llama,anil2023palm}. 
The power of LLMs only emerges when their parameter size exceeds a certain threshold \citep{wei2022emergent}, propelling the models to evolve in parameter scale.
Recent studies \citep{kaplan2020scaling,rae2021scaling,rosset2020turing} have demonstrated that larger models crave a massive, high-quality, and diverse pretraining corpus. 
The importance of data curation and assessment is increasingly evident.

\textbf{Data Curation:} 
Some work details preprocessing pipelines for specific sources like Common Crawl \citep{wenzek2019ccnet,abadji2022towards, penedo2023refinedweb} or Reddit \citep{gao2020pile}. 
However, these pipelines cannot be directly applied elsewhere because different curation pipelines should be built for various data sources by native speakers of target languages to ensure better quality control \citep{laurenccon2022bigscience}.
Unfortunately, an open-source system for customized pretraining data curation is still absent in the community.

\begin{figure} 
  \centering
  \includegraphics[width=0.48\textwidth]{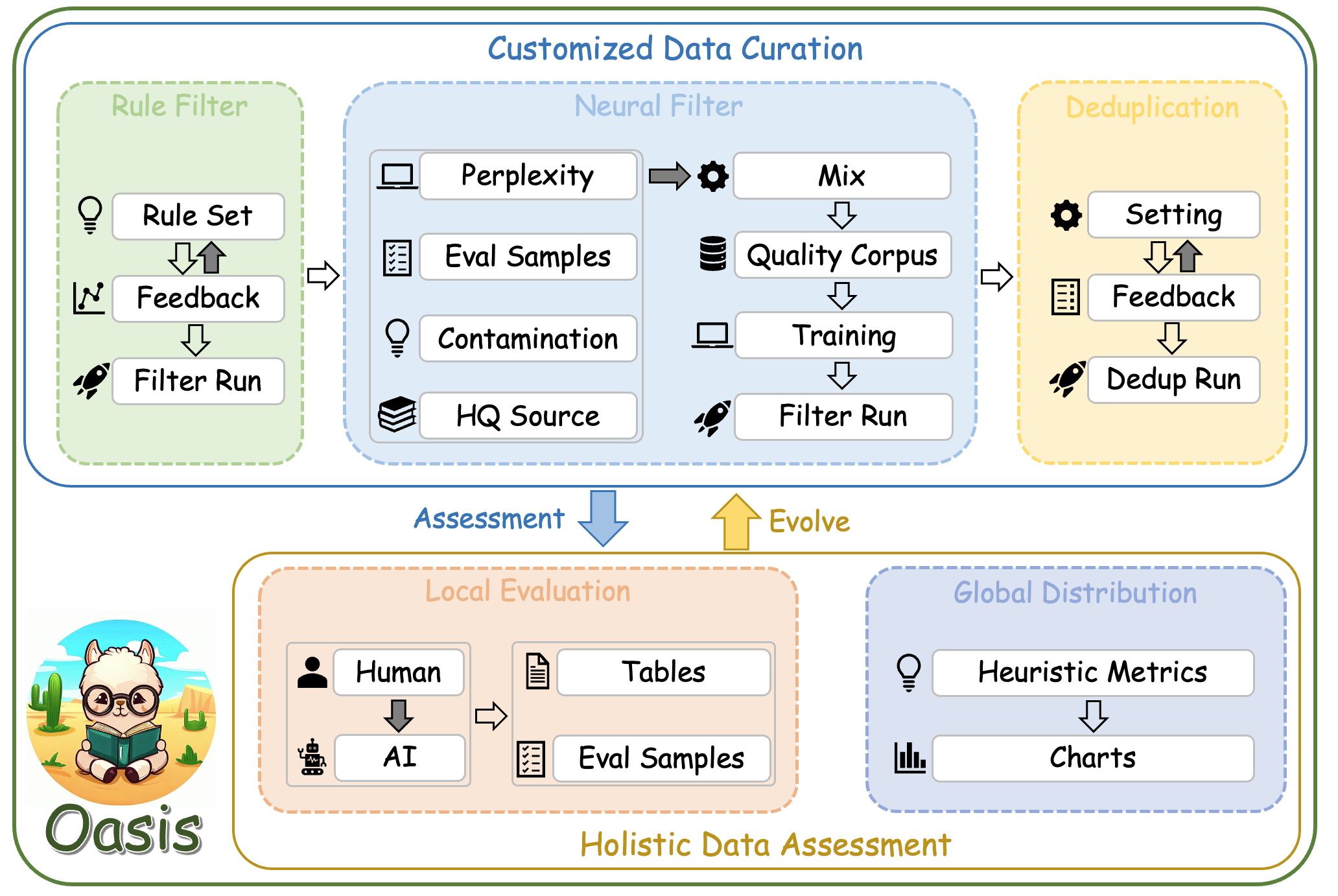}
  \caption{Overview of Oasis functionality.}
  \label{figure_task}
\end{figure}

\textbf{Data Assessment:} 
The assessment of the pretraining corpus \citep{kreutzer2022quality,dodge2021documenting} aids in the development of LLMs in a data-centric fashion \citep{fries2022bigbio} more efficiently.
It avoids optimizing the data curation by comparing the final model's performance after resource-consuming training.
Although there is no conclusion on quantifying the corpus's value, the consensus is that various aspects of pretraining data affect LLM performance, such as fluency, coherence, diversity, and bias \citep{longpre2023pretrainer,gunasekar2023textbooks}.
However, there is still a lack of a holistic data assessment system for the progressive improvement of the data curation pipeline.

In this paper, we present a system for customized pretraining data curation and holistic corpus assessment called \textbf{Oasis}. 
The functionality of this system covers three types of filters used to curate high-quality corpora and two perspectives for the holistic assessment of these corpora.

Specifically, in the \textbf{Customized Data Curation} part, the first step in our pipeline is an \textit{Interactive Modular Rule Filter} module, which enables users to construct the customized heuristic rule set with hit rate and bad cases as a reference.
Then, we debias the neural filter for text quality estimation by paying attention to the process of constructing source-specific quality classification datasets for training, constituting a \textit{Debiased Neural Filter} module.
Finally, in the \textit{Adaptive Document Deduplication} module, we optimize the widely used LSH deduplication method in memory requirement and exhibit the effect of different configurations for customized settings.
In the \textbf{Holistic Data Assessment} part, we provide options to inspect the corpus in sentence fluency and document coherence by humans or GPT-4 in the \textit{Local Quality Evaluation} module. The evaluated cases with quality labels could be further used to evolve the filtering pipeline.
Additionally, the \textit{Global Distribution Assessment} module displays the distribution information of the corpus in terms of diversity and richness by multiple heuristic metrics.

Aside from introducing Oasis, we demonstrate a complete case that utilizes this platform to build a high-quality and high-diversity Common Crawl corpus. 
Meanwhile, we holistic assess the corpus in the different development stages. The assessments also prove the effectiveness of the customized data curation process.
In addition, we publicly release an 800GB English-Chinese bilingual corpus Oasis-Corpus cultivated from web pages by Oasis to promote LLM development.

\section{Related Work}

\subsection{Data Cultivation}

The quantity \citep{hoffmann2022training} and quality \citep{gunasekar2023textbooks} of the pretraining corpus guaranteed LLM's performance in downstream tasks. 
State-of-the-art data cultivation methods can be classified into rule filter, neural filter, and deduplication.

\textbf{Rule filter}: \citep{penedo2023refinedweb,laurenccon2022bigscience,abadji2022towards} treat too low language identification confidence as a first criterion to drop the document. 
Moreover, some heuristic rules \citep{sun2021ernie,rae2021scaling,penedo2023refinedweb} focus on document length, punctuation ratio, word length, and stop words, which are widely used in deciding a document's quality.
\citep{laurenccon2022bigscience} also consider the closed class words ratio to distinguish machine-generated text.
Statistical language models like Kenlm \citep{heafield2011kenlm} are useful tools for efficiently estimating the coherence and fluency of sentences \citep{wenzek2019ccnet,laurenccon2022bigscience,wei2023polylm}. 
In removing undesirable information, \citep{gu2023eva2,wu2021yuan} build a word list to match and drop documents. \citep{rae2021scaling,wei2023polylm} utilize a URL block list to discard target web pages. 
While these methods could lead to bias, \citep{penedo2023refinedweb} optimize the block list by carefully reweighting these URLs.
However, the process of rule pipeline construction in a diverse customized corpus lacks attention.

\textbf{Neural filter}: Although the neural filter is more time-consuming than the rule filter, it can explore patterns between high- and low-quality data that cannot be literally concluded \citep{brown2020language}.
The training dataset utilizes well-known high-quality sources like Wikipedia, WebText \citep{radford2019language}, and Books as positive samples, meanwhile extensive various web pages as negative samples. 
A neural model like fastText or BERT trained on this dataset is responsible for scoring documents in quality \citep{touvron2023llama,brown2020language,gao2020pile}.
\citep{wu2021yuan} also consider utilizing a model to classify advertisements.
However, the neural filter could bias the filtered corpus due to the positive source of the training set \citep{dodge2021documenting,welbl2021challenges}.
Some works \citep{du2022glam,wei2023polylm} organize the positive sample in a mixture of various sources of high-quality texts to decrease the bias from the positive source.
\citep{penedo2023refinedweb} abandoned the neural filter on account of worrying about undesirable biases.

\textbf{Deduplication}: Repetition contents in pretraining corpus are proven to hurt the LLM's performance \citep{lee2021deduplicating}.
Corpus cultivation pipelines focus on fuzzy deduplication at the document \citep{zhang2022opt,biderman2023pythia,rae2021scaling} or line \citep{touvron2023llama} levels. 
These large-scale deduplication processes are mainly based on the locally sensitive hash algorithm \citep{rajaraman2011mining} by means of collision to calculate similarity.
\citep{sun2021ernie} calculate the MD5 of the three longest sentences to match the redundancy documents.
\citep{penedo2023refinedweb} further construct a huge prefix array to drop duplicate substrings.
These methods significantly improve the efficiency of the deduplication process, but the memory requirements become a barrier to deployment on a larger scale.

\subsection{Data Assessment}

Researchers have no consensus about the approaches in pretraining corpus assessment.
\citep{gao2020pile} visualize the various components of The Pile and utilize GPT-2 \citep{radford2019language} and GPT-3 \citep{brown2020language} to explore perplexity distribution and topic diversity. 
They also show the score distribution from the neural filter and the inspection of equality problems.
\citep{kreutzer2022quality} horizontally compare multiple corpora in linguistic correctness by human labeling.
\citep{luccioni2021s} focus on offensive content in the high- and low-quality scopes.
\citep{dodge2021documenting} explored the topic distribution of documents filtered by bad word lists and discovered harmless clusters like medicine and religion.
\citep{marone2023data} propose a more efficient algorithm for assessing data contamination in downstream tasks.
\citep{laurenccon2022bigscience} emphasize the difference among various languages, including the filtered ratio frequency distribution by different methods.
\citep{piktus2023roots,piktus2023gaia} build a tool to search strings in the entire corpus efficiently, providing the foundation for various exploration, like detecting personal identity information, inspecting undesired content, and fact verification.
There is still a lack of a holistic, multi-dimension, easy-to-use data assessment system.

\section{System Design and Algorithms}
In this section, we will introduce the system design of Oasis and detail the internal algorithms that differ from previous paradigms. 

\subsection{Customized Data Curation}
\subsubsection{Interactive Modular Rule Filter}
Building a rule filter for the pretraining corpus is a routine in state-of-the-art LLMs. A heuristic rule filter could preliminarily filter undesirable content efficiently. 
The heuristic ideas for building rules range from text length, punctuation, special tokens, blocklist, and language model perplexity. 
However, no rule sets can always be valid on various data sources and languages.
Corpora from different sources could vary in quality, style, format, template, and meta information. 
Filter rules in the book field may emphasize removing structural information among high-quality content. 
On the contrary, when handling documents from the massive web, rules would pay more attention to inspecting the content quality. 
The essential processes in building and improving the rules involve manually concluding patterns to distinguish high- and low-quality texts and adjusting a single heuristic by examining the hit samples.

We design functions in the Interactive Modular Rule Filter module according to the above intuitions.
A user builds a rule pipeline by interactively editing and connecting rule cells, referring to the patterns heuristic summarized from randomly displayed samples. 
A rule cell could be initiated with the predefined heuristic, and the user could also customize a heuristic function and add it to the predefined pool by typing Python code. 
Each rule cell's configuration, like thresholds and string patterns, can be freely adjusted according to the inspection of the hit rate and bad cases. 
After building a customized rule filter pipeline, Oasis can automatically generate a corresponding script according to settings and run the rule filter in the background.

\subsubsection{Debiased Model Filter}
The original intention of the neural filter is to select high-quality content from massive web pages, similar to high-quality sources like Wikipedia. 
The model can filter out content with non-summarizable patterns in quality aspects. 
However, treating another well-known high-quality source as positive and current sources as negative samples could lead the model to bias toward the high-quality source, affecting the quantity and diversity of the filtered data. 
\citep{penedo2023refinedweb} even abandoned this process due to scruples about the adverse effects of undesirable biases.

To address the bias issue, we propose a negative-centric dataset-building method for neural filter training. 
This method gathers the majority of positive samples from rule-filtered texts in the current source and obtains most negative samples through heuristic contamination of positive samples.
The predefined text contamination rule focuses on coherence and readability, involving shuffling, replacing, inserting, and deleting at the word, span, and sentence levels. 
The perplexities from the statistical language model may detect these undesirable low-quality contents. 
However, the perplexity metric is susceptible to low-frequency special tokens and biased towards the training corpus (usually Wikipedia). 
We use perplexity solely to identify extremely low-quality content, which constitutes a part of the negative samples.
These quality patterns are modeled using a neural filter with strong generalization capabilities, such as BERT. 
The finetuned BERT predicts scores for the text quality of every rule-filtered document.
We then drop documents according to the quality score below the threshold.

The Debiased Model Filter module provides a management panel for the quality classification dataset. 
Users can adjust the composition of positive and negative samples, customize text contamination rules based on editing feedback, and set perplexity quantiles to identify extremely low-quality content through case inspection.
Moreover, the dataset for neural classifier training could be further enhanced by incorporating evaluated texts from humans or GPT-4. 
After building a quality classification dataset, Oasis can generate corresponding scripts through parameter settings on the interface and run in the background with one click for neural filter training and the running process.

\begin{figure*} 
  \centering
  \includegraphics[width=0.93\textwidth]{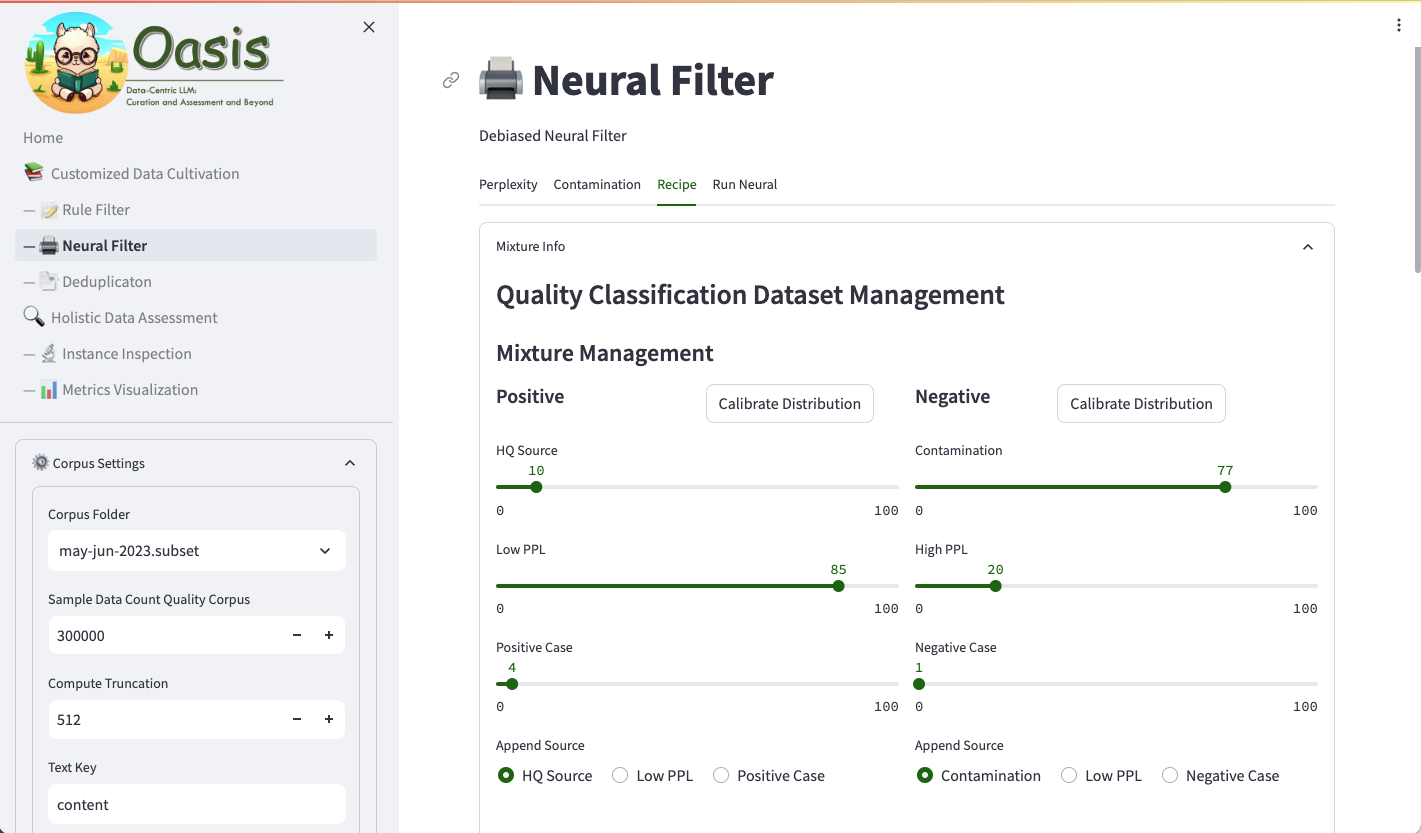}
  \caption{Screenshot of the recipe management interface of Oasis.}
  \label{figure_screenshot}
\end{figure*}

\subsubsection{Adaptive Document Deduplication}

Repetitive documents in the pretraining corpus would harm the LLM's generalization ability in various downstream tasks.
Massive deduplication among documents has a theoretical time complexity of $O(n^2)$.
The Locally Sensitive Hash algorithm approximates document similarity and reduces the time complexity, but it comes at the cost of increasing memory requirements to store hash collisions. 
Large-scale fuzzy deduplication becomes infeasible with limited resources.

\begin{small}
\begin{equation}
Pr(d_i, d_j | Jaccard(d_i, d_j)=s_(i,j)) = 1 - (1 - s^{b}_{i,j})^r 
\end{equation} % TODO: 过长换行
\end{small}

To achieve this goal, we reduce the memory requirement of the LSH deduplication algorithm to adapt to customized hardware by adjusting $r$ in the conditional probability formula. 
The system predicts the maximum $r$ according to the user's configuration in corpus size and memory size. 
Since a smaller $r$ will lead to a lower collision probability, the system also suggests the running times based on the Jaccard threshold and the expected duplication recall.

Although document-level deduplication could improve the diversity of the cultivated dataset, it could also significantly decrease the quantity. 
Our Adaptive Document Deduplication module also provides an interface to visualize the duplicated documents in a graph, offering options for users to make trade-offs between the removal rate and quantity.

\subsection{Holistic Data Assessment}

Evaluating LLMs pre-trained on different curated corpora using downstream tasks' performance serves as an oracle for assessing the data value. 
This post-hoc method is resource-consuming and ineffective. 
It is urgent to establish a holistic data assessment system to quantify the data quality and support the optimization process of data curation.
We achieve this goal through two views: local quality and global distribution, employing three evaluation methods: human assessment, heuristic metrics, and GPT-4.

\subsubsection{Local Quality Evaluation}
In this module, we focus on a document's fluency, readability, and coherence as assessed by humans or GPT-4. 
Due to the high consumption of the human inspection process, we only provide two quality options, "High" and "Low," in the user-friendly human evaluation interface. 
It displays real-time statistics of manually labeled quality conditions.
State-of-the-art (SOTA) LLMs like GPT-4 have demonstrated sufficient ability to score a document in multiple aspects, reflecting overall quality \citep{chen2023alpagasus}. % TODO：引用评估指令数据的工作 
We provide predefined prompts for quality assessment, achieving more than 95\% consistency with human opinions. 
The system also supports customized prompts for diverse demands. 
Moreover, the local quality evaluation samples can be incorporated into quality classification datasets to evolve the neural filter.

\subsubsection{Global Distribution Assessment}
Apart from the local document perspective, the global view of the corpus in statistical distribution can also reflect the broadly defined quality.

Oasis adopts six metrics to assess the corpus in heuristics from a randomly sampled subset of data:
(1) \textbf{Lexical Diversity Distribution} \citep{mccarthy2010mtld}: 
We calculate each document's Measure of Textual Lexical Diversity (MTLD) score to reflect lexical diversity and plot the frequency histogram to obtain an overall perspective.
(2) \textbf{Task2Vec Diversity Coefficient} \citep{lee2023beyond}: 
The task2vec diversity coefficient is proven to have a high correlation with humans' intuitive diversity of the corpus. We sample batches of text and display the calculated overall score.
(3) \textbf{Semantic Diversity Distribution}: 
We obtain all sampled documents' global semantic vectors using BERT and calculate the cosine similarity of each pair of documents to plot the frequency histogram.
(4) \textbf{Topic Diversity Distribution}: 
We cluster the sampled documents by global vector and calculate the similarity of centroid vectors among clusters to reflect overall topic diversity.
(5) \textbf{Knowledge Density and Diversity}: 
We inspect the knowledge view of the corpus by counting the different entities that occur. The density means the entities count normalized by word count, and diversity means the semantic similarity of all emerged entities.
(6) \textbf{Similarity to Wikipedia Distribution}: 
\citep{jansen2022perplexed} shows that the Kenlm model's perplexity on the target source could reflect the approximation of the Kenlm model's training source. We train a Kenlm model on Wikipedia and plot the perplexity distribution to inspect the extent of corpus bias in Wikipedia.

These metrics can be displayed on a single page and overlay multiple corpora for convenient visual comparison.

\begin{table*}[tp]\small
  \begin{center}
  \resizebox{1\textwidth}{!}{
    \begin{tabular}{l|cccc}
    \hline
        Corpus & Size & Human Rating & Knowledge Density & PPL in Wikipedia\\
    \hline
    WuDaoCorpus2.0-200G                     & 193 GB & 75\% & 7.11\% & 875.41 \\
    Oasis-Corpus-zh (with Debias Neural Filter)     & 370 GB & 90\% & 7.20\% & 922.97 \\
    Oasis-Corpus-zh (with Wiki-vs-CC Neural Filter) & $\sim$ 50 GB & 90\% & 7.99\% & 192.27 \\
    \hline
    \end{tabular}
}
    \caption{Comparison of evaluation metrics for different processing approaches on Chinese corpora. We obtain WuDaoCorpus2.0 from \citep{yuan2021wudaocorpora}. Oasis-Corpus-zh (with Wiki-vs-CC Neural Filter), has a data scale estimated based on the filter ratio.}
  \label{table_fix}
  \end{center}
\end{table*} 

\section{Usage Examples and Experiments}
In this section, we provide an example of how to interact with Oasis in data curation and assessment. We use the newest dump of Common Crawl (May/Jun 2023) as an illustration, focusing on English content.

\subsection{Customized Data Cultivation}
After the language identification and target language extraction pipeline \citep{abadji2021ungoliant} from WET files, we obtained a 2.4TB raw English dataset with meta information.

\textbf{Rule Filter:} Select the raw dataset in the Interactive Modular Rule Filter module and load the predefined rule pipeline. 
We can observe the hit rate for each rule cell and a random hit case after clicking the rule cell in the "Build Pipeline" panel.
Based on the case shown in the "Case Study" panel, an undesired advertising span is observed, inserted in a coherent sentence. 
Add a rule cell by setting arguments with the target span and clicking the "remove span" button in the left sidebar. 
Move up this cell before the last "min word count" cell. 
After saving the customized pipeline, you can find and load this pipeline in the configuration of the "Run Pipeline" panel and generate a runnable Python script for background multiprocessing running.
After applying this rule filter pipeline, we obtain 112GB of data.

\textbf{Neural Filter:} In the "Perplexity" panel, select a Kenlm model trained on Wikipedia and calculate perplexity to determine a quantile split between normal quality and extremely low-quality content. 
Drag the slider to change the quantile, inspect the cases, and finally decide on 0.85 as the boundary. 
Then, adjust the contamination set in the "Contamination" panel, following a logic similar to building a rule filter pipeline. 
In the "Recipe" section, manage the constitution of the quality classification dataset, both in positive and negative, build the dataset, and train a finetuned BERT model. 
Select the best checkpoint to run the neural filter in the "Run Neural" panel. 
Both the training and running processes occur in the background and do not affect other operations. 
After applying the neural filter, 100GB of high-quality data is obtained.

\textbf{Document Deduplication:} Utilize the duplication cluster graph to visualize the repeated pairs with different Jaccard thresholds in the "Dedup Case" panel. 
After a few trials, determine the Jaccard threshold as 0.8. 
In the "Run Dedup" panel, based on the corpus size and available memory, the system generates recommendations for parameter settings and can run document deduplication, utilizing multiple CPU cores in the background. 
The deduplication process finally removed 5\% of the documents.

\subsection{Holistic Data Assessment}

\textbf{Instance Inspection:} We aim to compare the quality of the corpus in its raw, rule-filtered, and neural-filtered states. 
First, select the four corpora and manually inspect each with 50 samples in the "Human Rating" panel. 
Set the default quality to high and click "low" only when a sample is not qualified for LLM training. 
The quality statistics are displayed in real-time in the sidebar. 
Then, in the "LLM Evaluation" panel, enter the API key for OpenAI and set 200 samples for each corpus to be evaluated, considering the cost. 
After receiving feedback for all the requests, check the average score of GPT-4. 
Conclusively, the document quality improves gradually as the build progresses.

\textbf{Heuristic Metrics:} 
In the "Heuristic Calculation" panel, select all heuristic metrics and choose multi-corpus for calculation. 
After obtaining the results for these metrics, pick the files to visualize in the "Report" panel. 
These charts demonstrate that our filter pipeline loses some diversity but substantially increases the quality. 
Our negative-centric dataset-building method introduced fewer biases than the previous wiki-vs-cc neural filters, achieving better lexical and topic diversity.

\subsection{Comparative Analysis}
As shown in Table \ref{table_fix}, the human-evaluated quality of the Chinese portion in the Oasis Corpus constructed by the Oasis system surpasses that of Wudao. 
Additionally, it exhibits a larger scale and greater knowledge diversity, demonstrating the advantage of Oasis, a comprehensive construction and evaluation system, over traditional data construction pipelines in pretraining data construction.

Compared to the corpora obtained by traditional positive-centric neural filters, the debias neural filter can produce comparable quality in human evaluation and a larger quantity.
The perplexities in the Wikipedia source also indicate that our neural filter could alleviate the bias toward high-quality sources in the corpus, ensuring diversity.

\section{Conclusion}
We propose Oasis, a one-stop system for LLM's pretraining data curation and assessment. 
In customized data curation, users can tailor their pipeline according to specific corpus requirements and limited hardware resources in rule filter, neural filter, and document deduplication. 
In holistic data assessment, a corpus can be evaluated from two perspectives: local document and global distribution; and in three ways: human assessment, GPT-4 evaluation, and heuristic metrics. 
These two components collaborate to enhance the value of the LLM's pretraining corpus.
The comparative analysis of the constructed corpora demonstrates the effectiveness of Oasis.

\section{Acknowledgements}
This work is supported by the National Key Research and Development Program of China (No. 2020AAA0106400), the National Natural Science Foundation of China (No. 61976211, 62176257).
This work is also supported by the Strategic Priority Research Program of Chinese Academy of Sciences (Grant No.XDA27020100), the Youth Innovation Promotion Association CAS, and Yunnan Provincial Major Science and Technology Special Plan Projects (No.202202AD080004)

% \bibliography{anthology,custom}
\bibliographystyle{acl_natbib}

\appendix

% \section{Appendix}
% \label{sec:appendix}

% This is a section in the appendix.

\end{document}